\documentclass{article}

     \usepackage[nonatbib,preprint]{neurips_2019}

\usepackage[utf8]{inputenc} 
\usepackage[T1]{fontenc}    
\usepackage{hyperref}       
\usepackage{url}            
\usepackage{booktabs}       
\usepackage{amsfonts}
\usepackage{amsmath}
\usepackage{amssymb}
\usepackage{nicefrac}       
\usepackage{microtype}      
\usepackage{graphicx}
\usepackage{url}
\usepackage{multirow}
\usepackage{url}
\usepackage{epsfig}
\usepackage{array}
\usepackage{times}
\usepackage{latexsym}
\usepackage{latexsym}
\usepackage{balance}
\usepackage{booktabs}
\usepackage{array,multirow,graphicx}
\usepackage{tikz}
\usepackage{amsmath}       

\usepackage{bm}
\usepackage[ruled]{algorithm2e}
\usetikzlibrary{shapes}
\usetikzlibrary{fit,positioning,calc,backgrounds}
\usetikzlibrary{decorations.pathreplacing}
\usepackage{array,multirow,graphicx}

\title{Wasserstein distances for evaluating cross-lingual embeddings}

\author{%
  Georgios Balikas \\
  Salesforce\\
  Grenoble, France\\
  \texttt{gbalikas@salesforce.com} \\
  \And
  Ioannis Partalas \\
  Expedia Group \\
  Geneva, Switzerland \\
  \texttt{ipartalas@expedia.com} \\
}

\begin{document}

\maketitle

\begin{abstract}
Word embeddings are high dimensional vector representations of words that capture their semantic similarity in the vector space. There exist several algorithms for learning such embeddings both for a single language as well as for several languages jointly.  In this work we propose  to evaluate collections of embeddings by adapting downstream natural language tasks to the optimal transport framework. We show how the family of Wasserstein distances can be used to solve cross-lingual document retrieval and the cross-lingual document classification problems. We argue on the advantages of this approach compared to more traditional evaluation methods of embeddings like bilingual lexical induction. Our experimental results suggest that using Wasserstein distances on these problems out-performs several strong baselines and performs on par with state-of-the-art models.  
\end{abstract}

\section{Introduction}
Cross-lingual word embeddings are vector representations of words from several languages in a shared, cross-lingual vector space. In this space, words from different languages with similar meanings obtain similar vectors. For example, the vectors of ``cat'' and ``chat'' (cat in French) are expected to be close. This property of cross-lingual embeddings (CLEs) to model the semantics of words irrespective of their language can power several cross-lingual applications and enable transfer learning from languages richer in resources to low resource ones \cite{ruder2017survey,agic2019jw300}.

From an application's perspective, CLEs promise is to enable cross-lingual NLP tasks like cross-lingual document retrieval or classification. This is done using transfer learning where one ``transfers'' knowledge from one language to another because of the common representation space.  This is done using a supervision signal in a language $\ell_1$ to ``learn'' the task, i.e., a function $\mathcal{F}$ that minimizes an appropriate loss. Then, given $\mathcal{F}$ and assuming the shared representation space between $\ell_1$ and $\ell_2$, inference can be performed in  $\ell_2$. This way one can transfer knowledge from $\ell_1$ to $\ell_2$. Consider, for instance,  e-commerce or travel industry  applications: one has data from user interactions on a variety of languages but it is costly and cumbersome to generate a labelled dataset for each supervised problem and language. On the other hand, if one would only require building training corpora  for $\ell_1$ in order to have a performing system across languages, it would greatly reduce the associated cost and human labour required. Hence, being able to transfer knowledge between languages efficiently in terms of both computational resources and measured performance is of great value.

The property of CLEs to model the meaning of words has been widely used for the evaluation of algorithms that learn CLEs. We argue, however, that lexicon induction is a task with limited application interest. Further, to the best of our knowledge there is no proof that lexicon induction performance is correlated with the performance on extrinsic tasks that are arguably the most interesting ones. In fact,  Glavaš et al. \cite{glavas2019properly} showed  empirically that the performance of CLE algorithms largely depends on the task. Their analysis also suggested  that optimizing CLE models for bilingual lexicon induction can result in poor  performance in downstream tasks. 
Here, we argue that complementary to the findings of \cite{glavas2019properly}, bilingual lexicon induction tasks evaluate CLEs by populating dictionaries with frequently used English words. For downstream tasks, however, it is not necessarily these words that guarantee satisfactory performance. In a classification task for example, one can obtain better performance by using the $N$-most important unigrams in terms of $\chi^2$  rather than using the $N$-most frequent unigrams \cite{manning2010introduction}. Thus, having access to a framework that benefits from high-quality embeddings and solves downstream tasks is a valuable evaluation resource.

Our goal in this paper is as follows. We propose a framework that evaluates  models for learning CLEs that builds on the optimal transport (OT). 
We argue that solving downstream NLP tasks using OT benefits from the quality of the embeddings. As a by-product of transforming the NLP applications to OT optimization problems, our method has very few free parameters. This enables reproducible results without requiring expensive hyper-parameter tuning iterations. In our experimental results, we demonstrate that this approach outperforms several baselines used frequently, and performs on par with state-of-the-art models explicitly proposed for the tasks.

\section{Optimal Transport for NLP}\label{sec:preliminary}
In this section we introduce the OT problem
\cite{kantorovich} and its entropic regularized version \cite{conf/nips/Cuturi13}. These versions of the OT problem will be later used to calculate document similarities using the Wasserstein distances. Having a way to calculate document distances using OT that capitalizes on the expressiveness of word embeddings, we will then evaluate families of word embeddings on a variety of downstream tasks. 

\subsection{Optimal transport} 
OT theory was first introduced in \cite{monge81} in order to study the problem of resource allocation. It provides a powerful geometrical tool that allows to compare probability distributions. 

In a more formal way, we assume access to two sets of points: the source points $X_S = \{\bm{x}^{S}_i \in \mathbb{R}^d\}_{i=1}^{N_S}$ and the target points $X_T = \{\bm{x}^{T}_i \in \mathbb{R}^d\}_{i=1}^{N_T}$. We then construct two discrete empirical probability distributions that model the source and target points as follows:
\[
\hat{\mu}_S = \sum_{i=1}^{N_S}p^S_i\delta_{\bm{x}_i^S} \text{ and } \hat{\mu}_T = \sum_{i=1}^{N_T}p^T_i\delta_{\bm{x}_i^T},
\]
where $p^S_i$ and $p^T_i$ are probabilities associated to $\bm{x}^{S}_i$ and $\bm{x}^{T}_i$, respectively and $\delta_{\bm{x}}$ is a Dirac measure that can be interpreted as an indicator function taking value 1 at the position of $\bm{x}$ and $0$ elsewhere. For these two distributions, the Monge-Kantorovich problem consists in finding a probabilistic coupling $\gamma$ defined as a joint probability measure over 
$X_S \times X_T$
with marginals $\hat{\mu}_S$ and $\hat{\mu}_T$ that minimizes the cost of transport with respect to some ground metric $l:X_s \times X_t \rightarrow \mathbb{R}^+$: 
\begin{equation}
\min_{\gamma \in \Pi(\hat{\mu}_S, \hat{\mu}_T)}\langle A, \gamma\rangle_F
\label{eq:wassersteinMinimization}
\end{equation}
where $\langle \cdot \text{,} \cdot \rangle_F$ is the Frobenius dot product, $\Pi(\hat{\mu}_S, \hat{\mu}_T) = \lbrace \gamma \in \mathbb{R}^{N_S \times N_T}_+ \vert \gamma \bm{1} = \bm{p}^S, \gamma^T \bm{1} = \bm{p}^T\rbrace$ is a set of doubly stochastic matrices and $A$ is a dissimilarity matrix, \textit{i.e.}, $A_{ij} = l(\bm{x}_i^S,\bm{x}_j^T)$ that defines the energy needed to move a probability mass from $\bm{x}_i^S$ to $\bm{x}_j^T$. This problem admits a unique solution $\gamma^*$ and defines a metric on the space of probability measures which is named  Wasserstein distance as follows:
$$W(\hat{\mu}_S, \hat{\mu}_T) = \min_{\gamma \in \Pi(\hat{\mu}_S, \hat{\mu}_T)}\langle A, \gamma\rangle_F.$$

Solving this problem relies on resource intensive algorithms. Recently, such algorithms have become popular in machine learning applications \cite{su2017order,balikas2018cross,kusner2015word,peyre2019computational} also due to the work of Cuturi et al.  \cite{conf/nips/Cuturi13}. They introduced an entropic regularized version of optimal transport that can be optimized efficiently using matrix scaling algorithm. We present this regularization below. 

\subsection{Entropic regularization}
The idea of using entropic regularization has recently found its application to the optimal transportation problem \cite{conf/nips/Cuturi13} through the following objective function:
\begin{equation}
    \min_{\gamma \in \Pi(\hat{\mu}_S,\hat{\mu}_T)}\langle A, \gamma\rangle_F - \frac{1}{\lambda}E(\gamma)
\label{eq:entropicRegulatization}
\end{equation}
The second term $E(\gamma) = -\sum_{i,j}^{N_S,N_T} \gamma_{i,j}\log(\gamma_{i,j})$ in this equation allows to obtain smoother and more numerically stable solutions compared to the original case and converges to it at the exponential rate \cite{2015-Benamou-Bregman}. The intuition behind it is that entropic regularization allows to transport the mass from one distribution to another more or less uniformly depending on the regularization parameter $\lambda$. Furthermore, it allows to solve the optimal transportation problem efficiently using Sinkhorn-Knopp matrix scaling algorithm \cite{sinknopp_67}. 

Notice that the only difference between Eq. \eqref{eq:wassersteinMinimization} and Eq. \eqref{eq:entropicRegulatization}
lies in the addition of the regularization term that offers smoother solutions to the problem. This suggests that using different regularization terms one can encode different properties in the solution of the minimization problem of Eq. \eqref{eq:wassersteinMinimization}. In this sense, entropic regularization allows for smoother and sparser solution. The formulation in both equations ignores the ordering of the elements. Recently, \cite{su2017order} extended the equations with an additional term to penalize plans according to the distance of elements in the source and target sequences. 


\subsection{Wasserstein distances for text mining applications}
In the framework of a text mining problem such as text classification, Kusner et al. \cite{kusner2015word} proposed to represent text as probability distributions using a bag-of-words representation. In their work they also proposed to calculate the dissimilarity matrix $A$ by taking advantage of word embeddings and using the Euclidean distance as a ground metric. Given that word embeddings can model the semantics of words in that semantically similar words are projected close in the embeddings space, using the Euclidean distance to capture the costs of transferring the words from the source text spans to the words of the target text spans is a natural choice. Therefore, when solving the minimization problem  of Eq. \eqref{eq:wassersteinMinimization}, words whose euclidean distance is small in the embeddings space will be transferred one to the other.   Instead of using bag-of-words representations as in \cite{kusner2015word}, in the rest of this work we opt for term-frequency inverse document frequency (tf-idf) representations that have been shown to perform well in IR tasks \cite{balikas2018ecir}. Also, in this work we will evaluate the effect of entropic regularization, that we expect to further improve the results, especially when the input documents are large.  

There are two main observations to be highlighted from the presentation of the OT measures above. First, the final distance between two documents requires solving an OT optimization problem whose inputs are the documents represented as discrete probability distributions and the per-word Euclidean distances dubbed $A$. The use of different families of embeddings impacts only the elements of $A$. Hence, we assume that more expressive $A$ due to higher quality embeddings should result in better performance in the text mining problems.

The second observation concerns the effect that hyper-parameter tuning can have on the obtained performance. When solving the optimization problems of Eq. \eqref{eq:wassersteinMinimization} and Eq. \eqref{eq:entropicRegulatization} one must select the values of \textit{zero} and \textit{one} hyper-parameter respectively. We emphasize that this is a very important advantage when it comes to evaluating different methods of word embeddings as the final performance does not require expensive fine-tuning.  Therefore, the only factor that impacts the performance on the downstream tasks is the embeddings used to populate $A$. We expect embeddings that learn accurate cross-lingual representations to perform better. Also, we expect embedding collections with larger embedding dictionaries to perform better.  These two hypotheses remain to be experimentally validated in the remaining sections of the paper.

We propose to follow a simple method for the tasks considered here, which is inspired by the Word Movers Distance of \cite{kusner2015word} and is based on nearest neighbors that leverage Wasserstein distances. Algorithm 1 presents the general method which can be applied to  both retrieval and classification tasks. In retrieval problems, given a ``query'' all documents are ranked in decreasing order of similarity with respect to the described Wasserstein distance. For classification, we use the labels of the top-$K$ most similar documents on a given document to decide its label.  Note that in the case of cross-lingual classification, we usually have labels in the source language which we use to classify documents in the target language. This framework is also known as zero-shot learning and is the one we will consider in the experimental sections. If labels are also available in the target language, one can combine the datasets in order to learn a classifier for two or more languages. The OT framework we proposed allows for that. Indeed, one could add labelled examples from the target language in the training data and evaluate their contribution. This is out of the scope of this paper though. 

\begin{algorithm}[H]
\SetAlgoLined
\KwData{$D_S$, $D_T$}
Generate $X_S, X_T$: the tf-idf representations of  $D_S, D_T$\;
\For{each \text{document} $i : d_i\in X_S$}{
\For {each \text{document} $ j : d_j \in X_T$}{
Calculate $A$, the Euclidean Distance (ground metric) between the elements of $d_i, d_j$\;
Calculate the $EMD(d_i, d_j)$ using Eq. \eqref{eq:wassersteinMinimization} or Eq. \eqref{eq:entropicRegulatization} \;
}
For $d_i$, return the  documents of $X_T$ in decreasing order of EMD distance\; 

}

\caption{Nearest Neighbors using Wasserstein Distance}
\end{algorithm}

\section{Evaluation Framework}
The goal of the evaluation section is to answer the following two questions: 
\begin{itemize}
    \item How do the nearest neighbors methods using the Wasserstein distances perform compared to other state-of-the-art moels and baselines? 
    \item What is the effect of different embeddings on the text mining problems considered here? 
\end{itemize}
To answer these questions we use two popular cross-lingual NLP tasks and compare the performance of three families of CLEs on these tasks. 

\subsection{Cross-lingual embeddings}

We continue our presentation with the CLEs we have used in this work. In order to ensure the replicability of our results we have used different cross-lingual, pre-trained embeddings  in the form they were released by the authors of the papers. While we present results using three families of embeddings due to space and computational limitations, any cross-lingual pre-trained embeddings can be used. Note that for all the embeddings we leverage their 300-dimensional versions.

\subsubsection{ConceptNet-Numberbatch} ConceptNet-Numberbatch embeddings \cite{speer2017conceptnetSemEval} are learned using the ConceptNet 5.5 knowledge base graph. The graph includes lexical and world knowledge from many different sources in several languages \cite{speer2017conceptnet}. To produce higher quality embeddings the methods also relies on expanding retrofitting. Expanded retroffiting \cite{speer2016ensemble} extends retrofitting \cite{faruqui2015retrofitting} and uses the graph and other available word embedding collections in order to learn embeddings for words that do not appear in the graph and improve the quality of the representations. Also, in the multilingual case it learns more about English words via their
translations in other languages, and also gives these foreign language terms useful embeddings in the same space as the
English terms.

\subsubsection{Smith et al.}
In their work Smith et al.  \cite{smith2017offline} where the authors demonstrate the benefits of using an orthogonal transformation between the spaces of a source and a target language. They show the benefits of using  singular value decomposition to obtain the transformation and an inverted softmax function in order to cope with the problem of hubness (a few words are the nearest neighbors of many words) when trying to find translations of words using k-Nearest neighbors \cite{dinu2014improving}. In our experiments these embeddings are dubbed \texttt{iclr} from the venue the work was presented.

\subsubsection{MUSE embeddings} 
In  \cite{conneau2017word} Conneau et al. proposed an unsupervised way to obtain cross-lingual embeddings in a shared vector space that relies on adversarial learning. The method assumes access to monolingual collections of embeddings and aligns them by projecting a source language to the shared space using a linear operation. Following \cite{smith2017offline} they enforce orthogonality and propose cross-domain similarity local scaling to overcome the hubness problem.

\subsection{Evaluation tasks}
We evaluate the performance of the different cross-lingual embeddings on two text mining tasks: i) cross-lingual sentence retrieval and ii) cross-lingual document classification. For both tasks we rely on the pre-trained embedding collections to embed words  in a shared space. Having the word representations, we compare how Wasserstein distances perform compared to other, commonly used methods that solve the same problems.   

We compare the performance of five different systems: 
\begin{itemize}
    \item \texttt{nBOW}: neural bag-of-words, where the word embeddings for the words of a sentence are averaged. To do that, we multiply the words' embeddings with the inverse document frequency (idf) of the word calculated in the available data. We normalize the resulting vector  so that its L$^2$ norm equals to 1.
    \item \texttt{EMD} This approach uses the Wasserstein distance described in Eq. \eqref{eq:wassersteinMinimization} to calculate distances between text spans. The smaller the distance, the closer the text spans semantically are. 
    \item \texttt{sEMD} This approach uses the regularized version of the Wasserstein distance described in Eq. \eqref{eq:entropicRegulatization} to calculate distances between text spans. Compared to  Eq. \eqref{eq:wassersteinMinimization} the expectation is that regularization can result in more robust solutions. 
    \item \texttt{ADAN} This is a deep neural method which implements a Deep Averaging Network structure and learns via an adversarial setting \cite{ADAN2018}.
    Concretely, the model learns a joint feature space for both languages while learning to classify the document in the source language and in the same time to discriminate the language of the document. The method is applicable on classification problems only.\footnote{We have used the implementation of the authors with the default parameters: \url{https://github.com/ccsasuke/adan}}
    \item \texttt{MLP-LASER}: This is an MLP classifier with two hidden layers which leverages the LASER sentence embeddings \cite{LASER}. We report the latest state-of-the-art results as described by the authors.\footnote{\url{https://github.com/facebookresearch/LASER/tree/master/tasks/mldoc}} Notice that the LASER sentence encoder is pre-trained using much more resources and data compared to our systems. 
\end{itemize}

For \texttt{nBOW}, \texttt{EMD} and \texttt{sEMD} we tuned the number of nearest neighbors using a validation set. We found that setting the value  of nearest neighbours to $K=5$ and the value of $\lambda=0.01$ performed well in the classification problems across languages and we have set them to these values for all datasets.  
For ADAN, we use the best settings as reported by the authors of \cite{ADAN2018}.

\subsubsection{Cross-lingual Sentence Retrieval}
For cross-lingual document retrieval we use the Europarl corpus \cite{koehn2005europarl}. It consists of the proceedings of the European Union Parliament in the form of sentences translated by humans in several European Union languages. Table \ref{tbl:sentenceRetrieval} presents the Precision at 1 (P@1) for different three pairs of languages: English with German (En-Ge), English with Italian (En-It) and English with Portuguese (En-Pt). For each language pair, we define two retrieval problems. For English with Italian for example, En$\rightarrow$It is the problem where the English sentences serve as queries and the Italian documents as the collection of sentences from where the translation sentence needs to be retrieved.         

From the P@1 scores of Table \ref{tbl:sentenceRetrieval} we observe that \texttt{sEMD} achieves the highest precision scores. This is the case independently of the language pair. The second best approach is \texttt{EMD}, which in most of the cases achieves comparable results. Both proposed methods obtain significantly higher scores compared to nBOW. nBOW performs poorly, suggesting that simply averaging word embeddings is not sufficient. 

In terms of embeddings we observe that the best performing embeddings are those released by Smith et al. as they achieve the highest performance in most of the language pairs. The second best performing family of embeddings are ConceptNet. An interesting observation from the results is the language of the query sentence is important wrt to the performance of the embeddings. 
For example, for En$\rightarrow$De ConceptNet achieves the best performance both for \texttt{EMD} and \texttt{sEMD}. For the case of De$\rightarrow$En however, Smith et al. perform better. We believe that this is due to training processes the authors of the embeddings followed and the cardinality of the vocabulary in each language. 

\begin{table}\setlength{\tabcolsep}{5pt}
    \centering
    \begin{tabular}{ll cc cc cc cc}
    \toprule
   Method  & Embeddings  & En$\rightarrow$De& De$\rightarrow$En& En$\rightarrow$It & It$\rightarrow$En& En$\rightarrow$Pt & Pt$\rightarrow$En \\  
\midrule

\multirow{3}{*}{nBOW} & conceptNet & 39.4 & 41.5 & 54.3 & 46.9 & 40.4 & 24.3 \\
 & muse & 42.2 & 32.2 & 38.3 & 28.8 & 42.5 & 34.0 \\
 & iclr & 48.6 & 35.8 & 57.4 & 44.6 & 60.2 & 49.1 \\\midrule

\multirow{3}{*}{EMD} & conceptNet & 86.4 & 82.9 & 90.7 & 90.4 & 78.4 & 81.2 \\
 & muse & 82.7 & 86.6 & 90.4 & 91.4 & 90.6 & 92.9 \\
 & iclr & 80.7 & 86.3 & 89.6 & 91.9 & 91.9 & 93.8 \\\midrule
\multirow{3}{*}{sEMD} & conceptNet & \textbf{86.8} & 84.1 & \textbf{91.3} & 90.8 & 79.3 & 81.1 \\
 & muse & 83.9 & \textbf{87.1} & 91.1 & 91.7 & 91.3 & 93.4 \\
 & iclr & 82.0 & \textbf{87.1} & 90.3 & \textbf{92.2} & \textbf{92.3} & \textbf{94.2} \\
     \bottomrule
    \end{tabular}
    \caption{Europarl Sentence  Retrieval; P@1 scores.}\label{tbl:sentenceRetrieval}
\end{table}




\subsubsection{Cross-lingual classification}
For the cross-lingual classification task we used the MLDoc corpus which is a balanced version of Reuters corpus \cite{mldoc2018}. The corpora consists of 1,000 training examples, 1,000 examples for validation and 4,000 examples for testing. We use the validation set for parameter tuning. The different versions based on Wasserstein distances are compared with ADAN and an MLP neural model that is based on the LASER sentence embeddings to represent the documents \cite{LASER}. Complementary to that, we dub EMD upper bound (``EMD UB'' in Table \ref{res:classification}) the performance of EMD with conceptNet embeddings when the training and test data are in the same language.  For this line of experiments \textit{only}, the reader should ignore the direction in the languages: En$\rightarrow$Fr assumes training data in French, and test data in French also.  This is why we note a single performance score in for  Fr$\rightarrow$En, $\ldots$ Es$\rightarrow$En: in these four cases the test data are in English and we use also the English training documents. 

Table \ref{res:classification} presents the results for cross-ligual classification in terms of accuracy along with the average rank of the methods across the different pairs of languages. We assign the ordinal ranks for each method in each pair of languages and we resolve ties by averaging the corresponding ranks. First, we observe that ADAN and MLP-LASER achieve the best results in 3 pairs of languages each followed by EMD which tops the scoreboard in two cases. Interestingly, in this case EMD is better than sEMD in most of the cases except the It$\rightarrow$En pair.


With respect to embeddings, we observe that ConceptNet are the best ones for both EMD and ADAN while the $iclr$ ones achieve the lowest scores across all languages and methods. Recall that in the retrieval task this family of embeddings achieved the best performance. Another interesting observation is that ADAN and EMD achieve better average rank with respect to the state-of-the-art MLP-LASER model which ranks third. Outperforming LASER which requires more training resources in terms of computational and data resources and is a system developed for such cross-lingual tasks shows the potential of OT for such applications.  


We now discuss the the performance of the systems that are using cross-lingual embeddings compared to the upper-bound of ``EMD UB''. We notice that the performance gap is lower when the training data are in English (En$\rightarrow$Fr, $\ldots$, En$\rightarrow$Es) compared to the cases where the training data are in the other four languages. This suggests that the CLEs we evaluate manage to transfer knowledge from English to other languages more efficiently than in the other way around.

\begin{table}\setlength{\tabcolsep}{5pt}
    \centering
    \begin{tabular}{ll cc cc cc cc r}
    \toprule
   Method  & Embeddings  & En$\rightarrow$Fr&  En$\rightarrow$It& En$\rightarrow$De & En$\rightarrow$Es& Fr$\rightarrow$En & It$\rightarrow$En & De$\rightarrow$En & Es$\rightarrow$En & Rank \\  
\midrule
EMD UB & cNet & 90.5 & 81.2 & 90.1 & 89.0 & \multicolumn{4}{c}{91.5} &  \\
\midrule
\multirow{2}{*}{EMD} & cNet & \textbf{86.5} & \textbf{74.2} & 86.9 & 76.3 & 79.0 & 70.2 & 76.6 & 70.0 & 2.6 \\
 & muse & 81.1 & 63.5 & 85.0 & 66.4 & 70.3 & 48.7 & 42.8 & 49.9 & 8.6 \\
 & iclr & 78.7 & 59.6 & 81.9 & 51.5 & 66.2 & 63.5 & 70.2 & 40.9 & 9.3 \\\midrule
 
\multirow{2}{*}{sEMD} & cNet & 84.8 & 71.7 & 86.0 & 73.4 & 76.7 & 72.3 & 76.0 & 72.3 & 3.6  \\
 & muse & 77.2 & 63.6 & 82.4 & 59.9 & 63.5 & 49.6 & 27.9 & 42.5 & 10.0 \\
 & iclr & 78.0 & 60.8 & 81.3 & 52.0 & 65.4 & 63.0 & 70.3 & 37.4 & 9.9 \\\midrule
\multirow{2}{*}{nBOW} & cNet & 74.5 & 65.7 & 80.7 & 73.5 & 76.3 & 69.0 & 77.9 & 67.8 & 6.7 \\
& muse & 77.1 & 63.0 & 77.3 & 67.4 & 73.1 & 63.1 & 65.2 & 69.2 & 8.6 \\
& iclr & 79.2 & 66.0 & 81.2 & 28.7 & 71.2 & 64.5 & 55.6 & 27.3 & 9.1 \\ \midrule
MLP & LASER & 78.0 & 70.2 & 86.2 & \textbf{79.3} & \textbf{80.1} & \textbf{74.1} & 80.7 & 69.6 & 2.8 \\ \midrule
\multirow{2}{*}{ADAN} & cNet & 82.3 & 70.3 & 86.0 & 75.6 & 80.0 & 72.3 & \textbf{80.8} & \textbf{76.4} & \textbf{2.5} \\
& muse & 78.0 & 66.9 & \textbf{87.7} & 74.6 & 72.0 & 65.6 & 77.6 & 64.5 & 5.3\\
& iclr & 59.8 & 46.9 & 69.3 & 57.7 & 61.5 & 49.5 & 58.8 & 43.0 & 11.6 \\

\bottomrule
\end{tabular}
\caption{Reuters Classification IDF. UB: upper bound. }
\label{res:classification}
\end{table}

\section{Related Work} Text mining tasks like text classification and information retrieval can be cast as distance calculation problems. Recently, Kusner et al. \cite{kusner2015word} proposed to estimate distances between text spans by using word embeddings to populate the ground metric matrix that is used by the  Wasserstein distance. The advantages of this approach are its superior performance, the lack of other free parameters that the ground metric matrix requires and its ability to model the geometry of the data. Following these observations, we show in this work how the family of Wasserstein distances with different forms of regularization can be used for evaluating CLEs on downstream cross-lingual tasks. We also propose simple approaches in the on-hand tasks that leverage Wasserstein distances. 

In a recent study the authors evaluate in an information retrieval task cross-lingual embeddings \cite{Litschko2019}. Apart from the fact that we also include a classification task in this paper we also focus on the inference part rather than relying on simple cosine similarities. 
OT methods for cross-lingual information retrieval has been presented in \cite{balikas2018ecir}. In this work we focus on the evaluation of the different CLE and we add another task that of classification. Additionally, we consider state-of-the-art deep neural  network approaches for the experimental comparisons.

\section{Conclusions}
In this work we presented an evaluation of multiple CLE in downstream tasks like cross-lingual retrieval and document classification. We propose a nearest neighbor approach based on Wasserstein distances that leverages CLE and compare it with state-of-the-art deep neural models. Our evaluation in the two NLP tasks shows that OT outperforms popular baselines for the task by a large margin and performs on par with other state-of-the-art systems for the task. Our findings open several avenues for future research. First, we would like to better understand the criteria under which regularization helps. In the problem of cross-lingual retrieval, we found that sEMD out-performed EMD but for classification we observed the opposite. From an application perspective for cross-lingual classification, it would be interesting to measure the benefit of adding documents in the same language. Our analysis showed an important opportunity when having access to data of the same language (the performance gap between EMD UB and EMD). How much one could gain when moving from zero-shot to few-shot learning for this problem needs to be estimated. Lastly, we showed that the OT methods performed on par with LASER, which is a state-of-the-art method for learning sentence embeddings. It would be interesting to apply EMD on the sentence level and compare its performance to our implementation here in order to measure how much information the sentence context can provide to the problem.

\bibliographystyle{plain}
\bibliography{biblio}

\begin{thebibliography}{10}

\bibitem{agic2019jw300}
{\v{Z}}eljko Agi{\'c} and Ivan Vuli{\'c}.
\newblock Jw300: A wide-coverage parallel corpus for low-resource languages.
\newblock In {\em Proceedings of the 57th Conference of the Association for
  Computational Linguistics}, pages 3204--3210, 2019.

\bibitem{LASER}
Mikel Artetxe and Holger Schwenk.
\newblock Massively multilingual sentence embeddings for zero-shot
  cross-lingual transfer and beyond.
\newblock {\em CoRR}, abs/1812.10464, 2018.

\bibitem{balikas2018ecir}
Georgios Balikas, Charlotte Laclau, Ievgen Redko, and Massih-Reza Amini.
\newblock Cross-lingual document retrieval using regularized wasserstein
  distance.
\newblock In {\em European Conference on Information Retrieval}, pages
  398--410. Springer, 2018.

\bibitem{2015-Benamou-Bregman}
Jean-David Benamou, Guillaume Carlier, Marco Cuturi, Luca Nenna, and Gabriel
  Peyr{\'e}.
\newblock {Iterative Bregman Projections for Regularized Transportation
  Problems}.
\newblock {\em {SIAM Journal on Scientific Computing}}, 2(37):A1111--A1138,
  2015.

\bibitem{ADAN2018}
Xilun Chen, Yu~Sun, Ben Athiwaratkun, Claire Cardie, and Kilian Weinberger.
\newblock Adversarial deep averaging networks for cross-lingual sentiment
  classification.
\newblock {\em Transactions of the Association for Computational Linguistics},
  6:557--570, 2018.

\bibitem{conneau2017word}
Alexis Conneau, Guillaume Lample, Marc'Aurelio Ranzato, Ludovic Denoyer, and
  Herv{\'e} J{\'e}gou.
\newblock Word translation without parallel data.
\newblock {\em arXiv preprint arXiv:1710.04087}, 2017.

\bibitem{conf/nips/Cuturi13}
Marco Cuturi.
\newblock Sinkhorn distances: Lightspeed computation of optimal transport.
\newblock In {\em NIPS.}, pages 2292--2300, 2013.

\bibitem{dinu2014improving}
Georgiana Dinu, Angeliki Lazaridou, and Marco Baroni.
\newblock Improving zero-shot learning by mitigating the hubness problem.
\newblock {\em arXiv preprint arXiv:1412.6568}, 2014.

\bibitem{faruqui2015retrofitting}
Manaal Faruqui, Jesse Dodge, Sujay~Kumar Jauhar, Chris Dyer, Eduard Hovy, and
  Noah~A Smith.
\newblock Retrofitting word vectors to semantic lexicons.
\newblock In {\em Proceedings of the 2015 Conference of the North American
  Chapter of the Association for Computational Linguistics: Human Language
  Technologies}, pages 1606--1615, 2015.

\bibitem{glavas2019properly}
Goran Glavas, Robert Litschko, Sebastian Ruder, and Ivan Vulic.
\newblock How to (properly) evaluate cross-lingual word embeddings: On strong
  baselines, comparative analyses, and some misconceptions.
\newblock {\em arXiv preprint arXiv:1902.00508}, 2019.

\bibitem{kantorovich}
Leonid Kantorovich.
\newblock On the translocation of masses.
\newblock In {\em C.R. (Doklady) Acad. Sci. URSS(N.S.)}, volume 37(10), pages
  199--201, 1942.

\bibitem{koehn2005europarl}
Philipp Koehn.
\newblock Europarl: A parallel corpus for statistical machine translation.
\newblock In {\em MT summit}, volume~5, pages 79--86. Citeseer, 2005.

\bibitem{kusner2015word}
Matt Kusner, Yu~Sun, Nicholas Kolkin, and Kilian Weinberger.
\newblock From word embeddings to document distances.
\newblock In {\em International Conference on Machine Learning}, pages
  957--966, 2015.

\bibitem{Litschko2019}
Robert Litschko, Goran Glava\v{s}, Ivan Vulic, and Laura Dietz.
\newblock Evaluating resource-lean cross-lingual embedding models in
  unsupervised retrieval.
\newblock In {\em Proceedings of the 42Nd International ACM SIGIR Conference on
  Research and Development in Information Retrieval}, SIGIR'19, pages
  1109--1112, 2019.

\bibitem{manning2010introduction}
Christopher Manning, Prabhakar Raghavan, and Hinrich Sch{\"u}tze.
\newblock Introduction to information retrieval.
\newblock {\em Natural Language Engineering}, 16(1):100--103, 2010.

\bibitem{monge81}
Gaspard Monge.
\newblock M\'emoire sur la th\'eorie des d\'eblais et des remblais.
\newblock {\em Histoire de l'Acad\'emie Royale des Sciences}, pages 666--704,
  1781.

\bibitem{peyre2019computational}
Gabriel Peyr{\'e}, Marco Cuturi, et~al.
\newblock Computational optimal transport.
\newblock {\em Foundations and Trends{\textregistered} in Machine Learning},
  11(5-6):355--607, 2019.

\bibitem{ruder2017survey}
Sebastian Ruder, Ivan Vuli{\'c}, and Anders S{\o}gaard.
\newblock A survey of cross-lingual word embedding models.
\newblock {\em arXiv preprint arXiv:1706.04902}, 2017.

\bibitem{mldoc2018}
Holger Schwenk and Xian Li.
\newblock A corpus for multilingual document classification in eight languages.
\newblock In {\em Proceedings of the Eleventh International Conference on
  Language Resources and Evaluation ({LREC}-2018)}, 2018.

\bibitem{sinknopp_67}
Richard Sinkhorn and Paul. Knopp.
\newblock Concerning nonnegative matrices and doubly stochastic matrices.
\newblock {\em Pacific Journal of Mathematics}, 21:343--348, 1967.

\bibitem{smith2017offline}
Samuel~L Smith, David~HP Turban, Steven Hamblin, and Nils~Y Hammerla.
\newblock Offline bilingual word vectors, orthogonal transformations and the
  inverted softmax.
\newblock {\em arXiv preprint arXiv:1702.03859}, 2017.

\bibitem{speer2017conceptnet}
Robyn Speer, Joshua Chin, and Catherine Havasi.
\newblock Conceptnet 5.5: An open multilingual graph of general knowledge.
\newblock In {\em 31st Conference on Artificial Intelligence}, pages
  4444--4451, 2017.

\bibitem{speer2017conceptnetSemEval}
Robyn Speer and Joanna Lowry-Duda.
\newblock Conceptnet at semeval-2017 task 2: Extending word embeddings with
  multilingual relational knowledge.
\newblock In {\em Proceedings of the 11th International Workshop on Semantic
  Evaluation (SemEval-2017)}, pages 85--89, 2017.

\bibitem{su2017order}
Bing Su and Gang Hua.
\newblock Order-preserving wasserstein distance for sequence matching.
\newblock In {\em Proceedings of the IEEE Conference on Computer Vision and
  Pattern Recognition}, pages 1049--1057, 2017.

\end{thebibliography}


\end{document}